\documentclass[letterpaper]{article}
\usepackage[preprint]{aaai2027}
\usepackage[hyphens]{url}
\usepackage{graphicx}
\usepackage{natbib}
\usepackage{caption}
\usepackage{subcaption}
\usepackage{amsmath,amssymb,bm,booktabs,multirow,array,xcolor}
\usepackage{tabularx}
\graphicspath{{figures/}}
\newcolumntype{P}[1]{>{\raggedright\arraybackslash}p{#1}}
\title{What Matters for Predicting the Next Tennis Shot? Local Context and Multi-Attribute Modeling}
\author{Qizhe Li\textsuperscript{1,2}, Junbiao Lu\textsuperscript{3}}
\affiliations{
\textsuperscript{1}The Chinese University of Hong Kong, Shenzhen\\
\textsuperscript{2}Shenzhen Research Institute of Big Data\\
\textsuperscript{3}City University of Hong Kong
}
\begin{document}
\maketitle

\begin{abstract}
Understanding how tennis players select their next shot is fundamental to characterizing their behavior during competitive matches. Each shot emerges from an evolving sequence of interactions, yet predicting a player's response from recorded match events remains challenging: the predictive value of preceding exchanges is not well understood, while shot attributes exhibit dependencies that vary across serving, returning, and rallying. In this work, we investigate how local interaction history informs multi-attribute next-shot prediction and how such dependencies can be modeled probabilistically. Through analysis of contextual information, we find that recent interactions provide substantial predictive value beyond the current playing state, with more benefits concentrated within a short history window. Building on this finding, we propose \textit{HITinnes}, a history-informed predictor that integrates interaction-aware event representations with phase-dependent conditional modeling. The framework further accommodates incomplete annotations through factor-wise supervision and produces target-level distributions via exact marginal inference. Experiments on professional men's-singles match records demonstrate that HITinnes outperforms statistical, neural, and alternative multi-attribute predictors, reducing Macro-NLL by 2.96\% relative to independent prediction heads while improving Macro-F1. Further analyses reveal the contributions of interaction representation, partial-label learning, and probabilistic inference, offering empirical insights into how temporal context and attribute dependencies shape the predictability of tennis shot behavior.
\end{abstract}


\section{Introduction}
\label{sec:introduction}

\begin{figure}[t]
\centering
\includegraphics[width=\columnwidth]{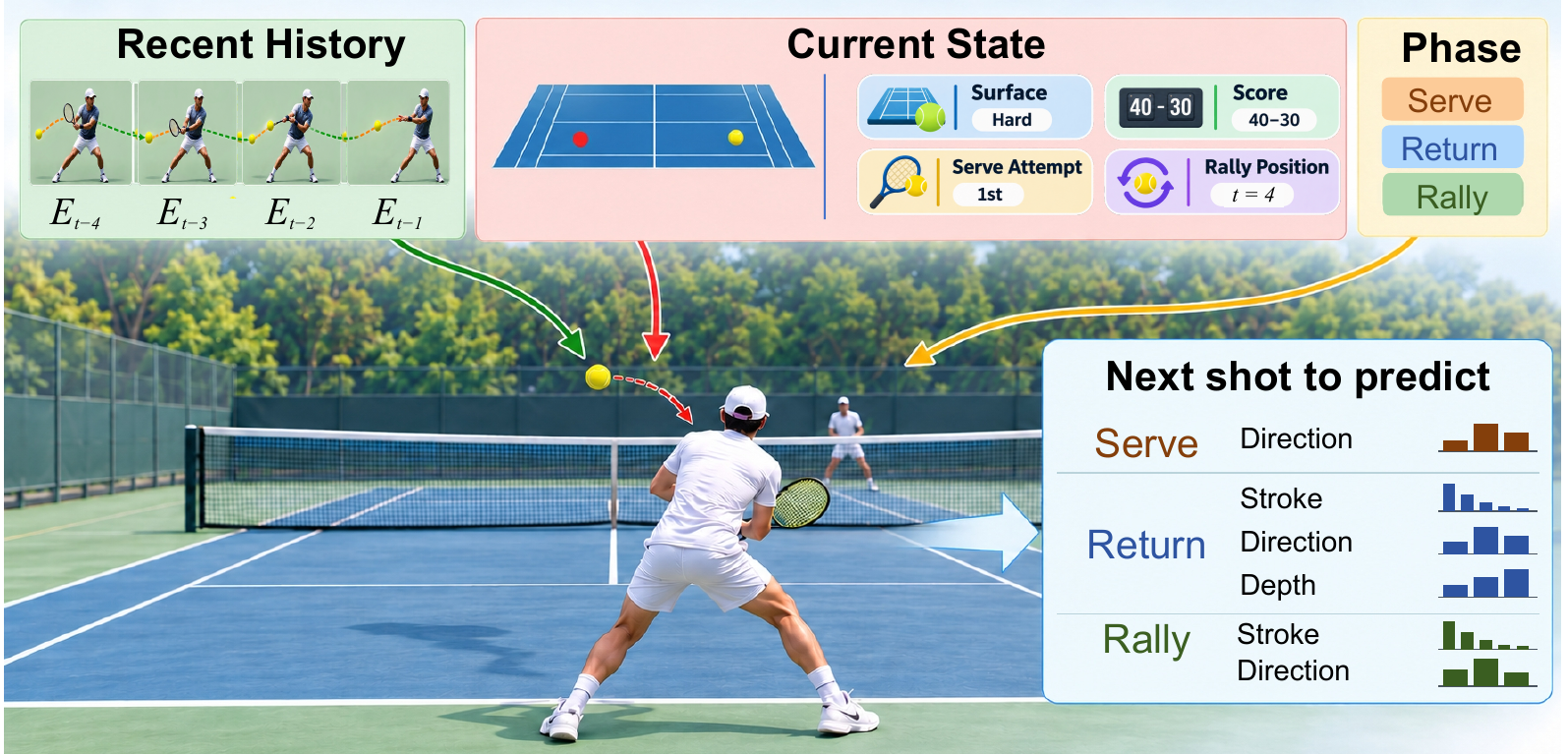}
\caption{Multi-attribute next-shot prediction in tennis. Given the pre-contact state, preceding interactions within the current point, and known playing phase, the predictor estimates distributions over the applicable shot attributes: direction for serves; stroke, direction, and depth for returns; and stroke and direction for rallies.}
\label{fig:overview}
\end{figure}

A tennis rally unfolds through a sequence of interactions, where each shot responds to preceding exchanges and shapes the context for the next action. Understanding and anticipating these shot choices is important for characterizing player behavior and identifying patterns in competitive play. Yet a player's next shot is not determined by the current situation alone: preceding interactions may reveal how the rally has developed, while multiple actions remain plausible before contact. This motivates a probabilistic view of next-shot prediction, in which the objective is to estimate the distributions of possible shot attributes from observable match information. As illustrated in Fig.~\ref{fig:overview}, the prediction targets vary across playing phases: serve prediction concerns direction; return prediction involves stroke type, direction, and depth; and rally prediction involves stroke type and direction. Such phase-dependent, multi-attribute prediction provides a finer characterization of shot-selection behavior than a single action label.

A fundamental question is what information makes the next shot predictable. Sequential event models have demonstrated the value of historical observations for forecasting subsequent events~\citep{mei2017neural,zhang2020selfattentive}. In racket sports, existing approaches incorporate rally progression, player characteristics, spatial information, and interaction history to predict future actions~\citep{fernando2020,wang2022shuttlenet,chang2023movement,pu2026}. However, these approaches differ in their available observations and prediction objectives, making it difficult to distinguish the contributions of individual contextual sources. In particular, how much recent shot interactions contribute beyond the current playing state, whether player identities provide additional information, and how far useful temporal dependencies extend remain important questions. Characterizing these contributions under a common observational setting can provide an empirical basis for selecting and representing the context used in next-shot prediction.

Beyond identifying informative context, a second challenge lies in modeling the attributes of the forthcoming shot. Stroke type, direction, and depth describe different aspects of an action, yet their distributions may be statistically dependent. Predicting each attribute independently overlooks these relationships, while treating every complete attribute combination as a distinct class directly parameterizes a larger joint output space. Conditional factorization offers an alternative representation of dependent categorical outputs through successive conditional distributions~\citep{read2009classifier,dembczynski2010bayes}. In tennis, this modeling problem is further complicated by phase-specific prediction targets and incomplete event annotations. An effective predictor must therefore connect historical context with a coherent distribution over the attributes applicable to each phase, while accommodating partially observed training labels and uncertainty in intermediate attribute predictions.

Together, these challenges raise a central research question: \textbf{How can local temporal dependencies be identified and used for structured probabilistic forecasting of the next tennis shot?} To address this question, we first conduct a controlled analysis of contextual information, examining the incremental predictive value of the current playing state, recent shot history, and player identities. History-window and lag-wise analyses further characterize the temporal scope of useful information. These findings guide the selection of a compact history representation for \textit{HITinnes}, a history-informed multi-attribute shot predictor. HITinnes encodes preceding events through interaction-aware features and temporal modeling, integrating their representations with the current pre-contact state. It then constructs phase-dependent joint distributions through conditional factors, supporting factor-wise learning from incomplete annotations and exact marginal inference over unknown upstream attributes.

We evaluate the proposed framework using professional men's-singles records from the Match Charting Project~\citep{sackmannMCP}. The contextual analysis reveals substantial predictive information in recent interactions beyond the current playing state, with most history-window gains captured within four preceding contacts. Under matched evaluation conditions, \textit{HITinnes} achieves improved multi-attribute prediction performance, reducing Macro-NLL by 2.96\% relative to independent prediction heads while also improving Macro-F1. Further comparisons and component analyses provide insights into the roles of interaction-aware representation and structured probabilistic modeling in next-shot prediction.

Our main contributions are summarized as follows:
\begin{itemize}
    \item \textbf{Predictive context characterization.}
    We systematically investigate the predictive contributions of the current playing state, local shot interactions, and player identities, characterizing the effective temporal scope and lag-wise sensitivity of historical information.

    \item \textbf{Structured probabilistic forecasting.}
    We develop \textit{HITinnes}, integrating interaction-aware history representations with phase-dependent conditional distributions, partial-label learning, and exact marginal inference for dependent shot attributes.

    \item \textbf{Empirical evaluation and insights.}
    We evaluate predictive performance on professional tennis records through comparisons with statistical, neural, and alternative multi-attribute predictors, together with controlled analyses of representation, supervision, and inference.
\end{itemize}

\section{Related Work}
\label{sec:related}

\paragraph{Sequential Modeling and Shot Prediction}
Sequential prediction models capture dependencies among preceding observations to anticipate subsequent events. Neural event models and attention-based architectures provide flexible mechanisms for representing such dependencies~\citep{mei2017neural,vaswani2017attention,zhang2020selfattentive}. In racket sports, memory-based models predict tennis shot locations from rally history~\citep{fernando2020}, Transformer-based approaches model tennis shot attributes~\citep{pu2026}, and ShuttleNet combines rally progression with player characteristics for badminton stroke forecasting~\citep{wang2022shuttlenet}. Further studies incorporate player interactions and spatiotemporal information for action and movement forecasting~\citep{chang2023movement,ibh2024stroke}. These approaches highlight the importance of interaction history, while differences in input information and prediction protocols motivate a controlled examination of which contextual observations contribute most to next-shot prediction.

\paragraph{Multi-Attribute Prediction and Tennis Analytics}
Predicting multiple related attributes can be approached through independent classifiers, joint categorical models, or conditional probability factorization. Autoregressive density estimation and probabilistic classifier chains establish ways to represent output dependencies~\citep{germain2015made,read2009classifier,dembczynski2010bayes}, while the relevance of these dependencies depends on the prediction objective~\citep{dembczynski2012dependence}. In tennis, notational analysis and event-level datasets support the study of shot patterns and player behavior~\citep{martinez2018tactical,torresluque2018,terroba2010tactical,sackmannMCP,zhao2025}. Complementary approaches use spatial tracking and tactical visualization to analyze interactions and playing strategies~\citep{wei2013,kovalchik2020voncramm,wu2021tacticflow,wu2022rasipam}. In table tennis, video-based technique recognition combines visual observations with contextual knowledge~\citep{he2024vistec}. \textit{HITinnes} addresses phase-dependent, multi-attribute prediction from categorical shot records, enabling us to examine how conditional modeling of attribute relationships complements informative temporal context.

\section{Problem Formulation}
\label{subsec:problem}

\paragraph{Prediction Setting and Context}
We consider a tennis point as an ordered sequence of shot events $(E_0,\ldots,E_T)$, where $T$ is the final contact index. At each contact $t$, the task is to predict the forthcoming shot using only information available before that contact.

Let $S_t$ denote the observed pre-contact state, including surface, point/game/set score, serve attempt, rally position, and serving role, and let $\phi_t\in\{\mathrm{serve},\mathrm{return},\mathrm{rally}\}$ denote the known shot phase. The preceding $k$ events form a local history
\[
H_t^{(k)}=(E_{\max(0,t-k)},\ldots,E_{t-1}),
\]
which is reset at each point boundary. Acting-player and opponent identities, denoted by $P_t$ and $O_t$, are also considered as candidate contextual information. The available context is therefore $X_t^{(k)}=(S_t,H_t^{(k)},P_t,O_t,\phi_t)$, from which different input configurations are evaluated. All current-shot attributes and subsequent outcomes are excluded from the prediction inputs.

\paragraph{Phase-Dependent Prediction Targets}
The next shot is represented by multiple categorical attributes whose applicability depends on the current phase. Let $R_t$, $D_t$, and $Z_t$ denote stroke type, direction, and return depth, respectively. The phase-specific action is defined as
\begin{equation}
A_t=
\begin{cases}
D_t^{\mathrm{serve}}, & \phi_t=\mathrm{serve},\\
(R_t,D_t,Z_t), & \phi_t=\mathrm{return},\\
(R_t,D_t), & \phi_t=\mathrm{rally}.
\end{cases}
\label{eq:actions}
\end{equation}
The attributes are categorical with phase-specific vocabularies, yielding six prediction targets across the three phases. Some applicable attributes may be unannotated, and such missing labels are excluded from the corresponding supervised objectives and evaluations through the corresponding observation masks.

Let $C_t$ denote an input configuration formed by selecting components of the candidate context $X_t^{(k)}$, and let $\theta$ denote the trainable model parameters. Our goal is to estimate the conditional distribution $p_\theta(A_t\mid C_t)$ and the associated marginal distributions of its attributes. \textit{HITinnes} uses $C_t=(S_t,H_t^{(k^\star)},\phi_t)$, where $k^\star$ is selected through temporal-context analysis. The resulting model characterizes the distribution of observed shot choices conditioned on the available pre-contact information.

\section{Predictive Value of Shot Context}
\label{sec:preliminary}

Before developing the prediction model, we investigate how observable context contributes to next-shot prediction. Our study examines the incremental value of different information sources, the effective temporal scope of shot history, and prediction sensitivity to individual historical events.

\paragraph{Incremental Value of Context}
We first compare four nested input configurations under matched model capacity and training settings. Starting from the pre-contact state ($M_S$), we progressively incorporate recent shot history ($M_H$), acting-player identity ($M_P$), and opponent identity ($M_{PO}$). The incremental contribution of each information source is measured by the reduction in Macro-NLL, which equally averages the negative log-likelihood across six phase-specific prediction targets. Identity-specific residual effects are regularized by training support using $g(n,\tau)=n/(n+\tau)$, where $n$ is the corresponding player or opponent's training-contact count. We use $\tau=5000$ for both identity residuals.

\begin{figure}[t]
    \centering
    \begin{subfigure}[t]{0.45\columnwidth}
        \includegraphics[width=\linewidth]
        {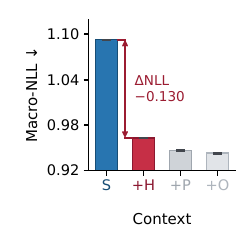}
        \caption{Context gains.}
    \end{subfigure}\hfill
    \begin{subfigure}[t]{0.45\columnwidth}
        \includegraphics[width=\linewidth]
        {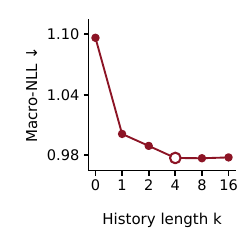}
        \caption{History window.}
    \end{subfigure}
    \caption{Predictive value of local context.}
    \label{fig:rq1-results}
\end{figure}

In Fig.~\ref{fig:rq1-results}(a), S, +H, +P, and +O denote the successive state, history, player, and opponent configurations. Error bars show one sample standard deviation (SD) across three seeds. Recent history provides the largest incremental improvement. Adding preceding contacts reduces Macro-NLL from 1.0928 to 0.9631, corresponding to an 11.9\% relative reduction. Incorporating player and opponent identities yields smaller additional improvements, reaching 0.9465 and 0.9427, respectively. These results indicate that recent interactions provide substantial predictive information beyond the current state under the evaluated context configurations. Accordingly, \textit{HITinnes} uses the current state and local shot history as its prediction context.

\paragraph{Effective Temporal Scope}
We next examine how predictive performance changes as the historical context is extended. Using a development subset drawn exclusively from the training partition, we evaluate a sequence of candidate history lengths while keeping the prediction targets and model configuration fixed. As shown in Fig.~\ref{fig:rq1-results}(b), Macro-NLL decreases sharply when the first few preceding contacts are introduced, while additional history yields diminishing improvements. The predictive benefit largely saturates around four preceding contacts.

We therefore select $k^\star=4$ as a compact operating horizon for \textit{HITinnes}. This selection is based on the development-subset screen, independently of the formal held-out test matches. The resulting horizon balances the predictive value of recent interactions against the inclusion of increasingly distant events.

\paragraph{Sensitivity to Historical Lags}
To further characterize the temporal distribution of predictive information, we examine how a fitted history model responds to perturbations at different lag positions. Lag occlusion removes a selected historical event and compacts the remaining sequence, whereas conditional permutation exchanges complete event tokens among samples matched by prediction phase and lag-specific actor role. Both analyses keep the model parameters fixed and measure the resulting increase in Macro-NLL.

Prediction is considerably more sensitive to recent interactions. Permuting the immediately preceding event increases Macro-NLL by 0.2837 nats, compared with 0.0015 nats at lag four. Occlusion exhibits the same ordering, with increases of 0.2779 and 0.0055 nats, respectively. Occlusion also changes the positions of retained events. The consistent sensitivity ordering across both perturbations supports the importance of the nearest interaction.

Together, these findings characterize localized temporal dependence in the evaluated shot sequences. A compact history captures most of the observed predictive gain, and the fitted model is most sensitive to the nearest contacts. These observations motivate the history-conditioned representation developed in the following section.

\section{History-Informed Shot Prediction with \textit{HITinnes}}
\label{sec:method}

\begin{figure*}[t]
    \centering
    \includegraphics[width=\textwidth]
    {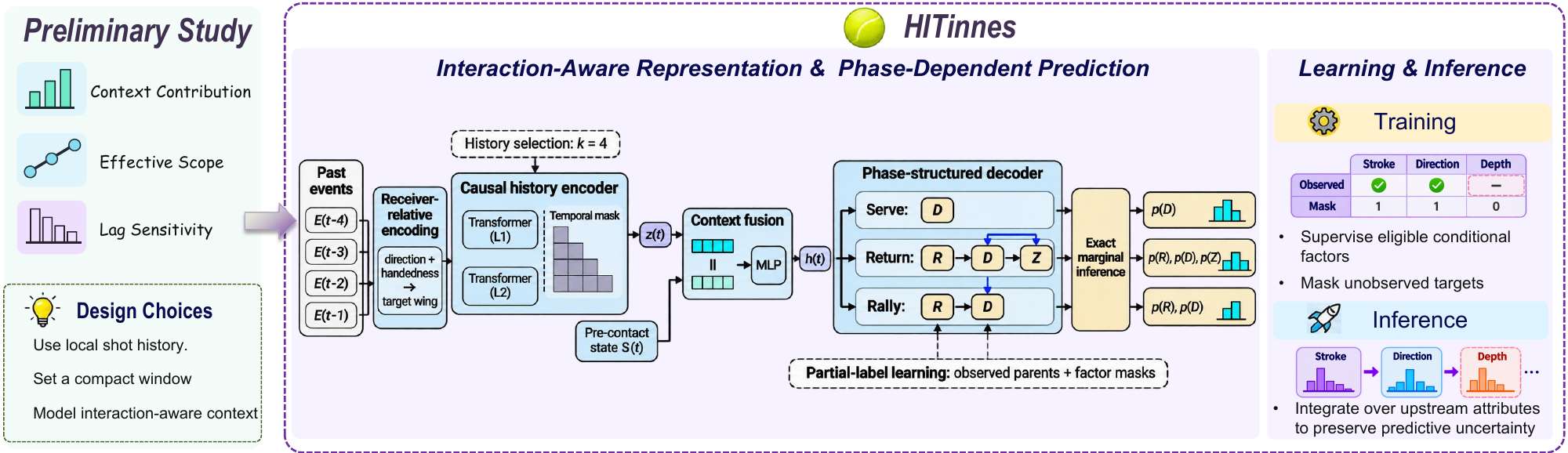}
    \caption{\textit{HITinnes} framework. The preliminary study selects the history horizon. Receiver-relative event encoding and a causal sequence encoder summarize the retained contacts, which are fused with the pre-contact state. Phase-specific conditional factors support partial-label learning and exact marginal prediction.}
    \label{fig:framework}
\end{figure*}

Motivated by the predictive value of local interactions identified in Section~\ref{sec:preliminary}, we develop \textit{HITinnes}, a \textbf{H}istory-\textbf{I}nformed model for multi-attribute tennis shot prediction. \textit{HITinnes} encodes preceding shot interactions together with the current pre-contact state and models the next shot through phase-dependent conditional distributions. As illustrated in Fig.~\ref{fig:framework}, the framework integrates interaction-aware history representation, conditional action modeling, and probabilistic learning and inference.

\subsection{Interaction-Aware History Representation}
\label{subsec:representation}

\textit{HITinnes} summarizes the preceding $k^\star$ contacts into a representation of local shot interactions. While the original event records describe individual shot attributes, their interpretation can depend on the relationship between the acting player and the receiver. We explicitly encode this relationship before aggregating the historical events.

\paragraph{Receiver-Relative Encoding}
We preserve the charted direction $d_i$ and derive its receiver-relative target wing using the mapping $\rho$.
\begin{equation}
w_i=\rho(d_i,q_i),\quad
w_i\in\{\mathrm{FH},\mathrm{BH},
\mathrm{middle},\mathrm{unknown}\},
\label{eq:wing-transform}
\end{equation}
where $q_i$ is receiver handedness, and FH and BH denote forehand and backhand. For lateral direction code 1, the mapping yields FH for a right-handed receiver and BH for a left-handed receiver. Code 3 reverses these assignments. Code 2 maps to middle regardless of handedness. Serve-specific codes and unresolved combinations of direction and handedness map to unknown. Their original direction remains available through $d_i$, while $\rho$ adds receiver-relative categorical semantics.

For each preceding event, $x_i$ contains stroke, depth, phase, serve number, and six charted modifiers comprising approach, serve-and-volley, court position, net-cord contact, drop-volley indication, and fault type. We concatenate their embeddings with those of $d_i$, $w_i$, and actor role $r_i$, which indicates whether the historical actor is the current target player or opponent. A learned projection $W_e$ maps these feature embeddings into a shared event space.
\begin{equation}
\begin{aligned}
e_i=W_e[&
\operatorname{Emb}(x_i)\Vert
\operatorname{Emb}(d_i)\Vert\\
&\operatorname{Emb}(w_i)\Vert
\operatorname{Emb}(r_i)]+p_i,
\end{aligned}
\label{eq:event-representation}
\end{equation}
where $\operatorname{Emb}$ denotes learned categorical embeddings, $\Vert$ denotes concatenation, $e_i\in\mathbb R^{d_e}$, and $p_i$ is a learned positional embedding within the retained history. The resulting event token combines the recorded action, its interaction-relative semantics, and its position in the local sequence.

\paragraph{Temporal Encoding and Context Fusion}
A causally masked Transformer maps the retained within-point contacts to a history summary $z_t\in\mathbb R^{d_e}$. Context fusion combines this summary with the current state to produce $h_t\in\mathbb R^{d_h}$. The dimensions and sequence-mask implementation are specified in Section~\ref{subsec:setup}.

Let $\ell_t=\max(0,t-k^\star)$. The history encoder $\operatorname{Enc}_{\mathrm{hist}}$ produces a summary that is fused with the current pre-contact state.
\begin{equation}
\begin{aligned}
z_t&=\operatorname{LastValid}\Big(
\operatorname{Enc}_{\mathrm{hist}}
(\texttt{START},e_{\ell_t},\ldots,e_{t-1})
\Big),\\
h_t&=F_\theta\big(
\operatorname{Emb}(S_t,\phi_t),z_t
\big),
\end{aligned}
\label{eq:context-encoding}
\end{equation}
where $\operatorname{LastValid}$ extracts the final valid encoded position and $F_\theta$ denotes the context-fusion network. For the first contact, the START token supplies the historical representation.

The fused vector $h_t$ combines the current playing context with preceding shot interactions. It is shared across all prediction targets, providing a common representation from which the phase-dependent action distribution is constructed. The target-wing configuration also embeds acting-player and opponent handedness in the current state. The pressure variant additionally concatenates serve side, game-pressure category, set-pressure category, and a tiebreak indicator into this state embedding before $F_\theta$. Game pressure distinguishes ordinary play, deuce, and server/receiver game-point or tiebreak-point states. Set pressure identifies whether either player can close the set. These features are computed from pre-contact scores.

\subsection{Phase-Dependent Conditional Prediction}
\label{subsec:prediction}

Given $h_t$, \textit{HITinnes} estimates a distribution over the action attributes applicable to the current phase. Independent prediction heads model each attribute separately, while a flat joint classifier assigns probabilities to complete action combinations. We instead use conditional factorization to represent dependencies among attributes through reusable conditional prediction components.

\paragraph{Conditional Action Factorization}
Following the action definition in Eq.~\ref{eq:actions}, the serve phase is represented by a direction distribution. Let $r$, $d$, and $z$ denote candidate stroke, direction, and depth categories. The return and rally joint distributions are factorized as
\begin{equation}
\begin{aligned}
p_\theta^{\mathrm{ret}}(r,d,z\mid h_t)
&=p_\theta^{\mathrm{ret}}(r\mid h_t)
p_\theta^{\mathrm{ret}}(d\mid r,h_t)\\
&\quad\cdot p_\theta^{\mathrm{ret}}(z\mid r,d,h_t),\\
p_\theta^{\mathrm{ral}}(r,d\mid h_t)
&=p_\theta^{\mathrm{ral}}(r\mid h_t)
p_\theta^{\mathrm{ral}}(d\mid r,h_t).
\end{aligned}
\label{eq:action-factorization}
\end{equation}
Serve direction is modeled directly as
$p_\theta^{\mathrm{ser}}(d\mid h_t)$.
Here $\mathrm{ret}$, $\mathrm{ral}$, and $\mathrm{ser}$ denote return, rally, and serve, respectively.

For return, the stroke distribution is conditioned on the contextual representation, direction additionally depends on stroke, and depth depends on both stroke and direction. Rally follows a corresponding two-attribute factorization. This ordering specifies statistical dependence within a single action. The attribute order is a modeling factorization of the contemporaneous shot attributes.

Each factor is implemented as a categorical prediction head. The first factor receives $h_t$, whereas downstream factors additionally receive embeddings of their upstream attributes. For example, the direction logits are
\begin{equation}
\ell_D^\phi(r,h_t)
=W_D^\phi[h_t\Vert v_R^\phi(r)]+b_D^\phi,
\label{eq:conditional-heads}
\end{equation}
for $\phi\in\{\mathrm{return},\mathrm{rally}\}$, where $r$ is a stroke category, $\ell_D^\phi$ is the vector of direction logits, $W_D^\phi$ and $b_D^\phi$ are learned parameters, and $v_R^\phi(r)$ is the phase-specific stroke embedding. The return-depth factor analogously receives embeddings of both stroke and direction.

Applying softmax to each head yields normalized conditional probabilities. Their product defines a normalized joint distribution for each phase. The conditional heads represent relationships among output attributes and share prediction components across action combinations.

\subsection{Partial-Label Learning and Marginal Inference}
\label{subsec:learning}

The conditional structure supports different uses of attribute information during training and prediction. Observed upstream attributes provide conditioning values for eligible factors during learning. At inference, the model integrates uncertainty over possible upstream values through its conditional distribution.

\paragraph{Partial-Label Supervision}
The event records contain partially annotated shots. \textit{HITinnes} uses the available labels to supervise eligible conditional factors. A factor contributes to the training objective when both its target and the upstream attributes on which it conditions are annotated.

Let $m_{t,j}$ indicate whether target $j$ is applicable and observed at contact $t$, and let $\mathcal P(j)$ denote the upstream attributes required by its conditional factor. The effective supervision mask is
\begin{equation}
\widetilde m_{t,j}
=m_{t,j}\prod_{q\in\mathcal P(j)}m_{t,q}.
\label{eq:training-mask}
\end{equation}
For example, a return with observed stroke and direction but missing depth can supervise its first two factors. The direction factor receives supervision when both direction and its conditioning stroke are observed. For factors with no upstream attributes, the empty product in Eq.~\ref{eq:training-mask} equals one.

For a minibatch $\mathcal B$, define
$N_{\mathcal B,j}=\sum_{t\in\mathcal B}\widetilde m_{t,j}$
and
$\mathcal J_{\mathcal B}=\{j\mid N_{\mathcal B,j}>0\}$. Here $N_{\mathcal B,j}$ counts eligible observations for target $j$, and $\mathcal J_{\mathcal B}$ contains the targets with eligible minibatch labels.
The target-balanced conditional negative log-likelihood is
\begin{equation}
\begin{aligned}
\mathcal L_{\mathcal B}(\theta)
=&\frac{1}{|\mathcal J_{\mathcal B}|}
\sum_{j\in\mathcal J_{\mathcal B}}
\frac{1}{N_{\mathcal B,j}}
\sum_{t\in\mathcal B}\widetilde m_{t,j}\\
&\quad\cdot
\left[-\log p_\theta^{\phi_t}
(a_{t,j}\mid h_t,a_{t,\mathcal P(j)})
\right].
\end{aligned}
\label{eq:loss}
\end{equation}
Here $a_{t,j}$ is the observed value of target $j$, and $a_{t,\mathcal P(j)}$ collects its observed upstream values. The objective balances eligible conditional likelihoods across the available targets and preserves factor-wise supervision from partially annotated events.

\paragraph{Exact Marginal Inference}
At prediction time, \textit{HITinnes} propagates upstream uncertainty into each target distribution through exact marginalization of the phase-specific joint distribution.
\begin{equation}
\bar p_{t,j}(a)=
\sum_{a_{t,-j}}
p_\theta^{\phi_t}
(a_{t,j}=a,a_{t,-j}\mid h_t),
\label{eq:marginal-prediction}
\end{equation}
where $a$ is a candidate value of target $j$, and $a_{t,-j}$ denotes the remaining phase-applicable attributes.

For return depth, the marginal sums over all stroke and direction combinations for each depth category $z$.
\begin{equation}
\begin{aligned}
\bar p_{t,Z}(z)
=&\sum_{r,d}
p_\theta^{\mathrm{ret}}(r\mid h_t)
p_\theta^{\mathrm{ret}}(d\mid r,h_t)\\
&\quad\cdot p_\theta^{\mathrm{ret}}(z\mid r,d,h_t).
\end{aligned}
\label{eq:depth-marginal}
\end{equation}
Since the action spaces are finite and categorical, the required sums can be evaluated exactly. This produces normalized marginal probabilities for the six phase-specific targets while preserving uncertainty across dependent action attributes.

\section{Experiments}
\label{sec:experiments}

\subsection{Research Questions}
\label{subsec:research-questions}

We evaluate \textit{HITinnes} through two research questions, focusing on predictive effectiveness and the contributions of its key modeling choices.

\begin{itemize}

\item \textbf{RQ1: Prediction Effectiveness.}
How effectively does \textit{HITinnes} predict multi-attribute shot distributions compared with existing statistical, neural, and structured prediction approaches?

    \item \textbf{RQ2: Design Analysis.}
    How do interaction-aware representation, partial-label supervision, and probabilistic inference contribute to prediction quality?
\end{itemize}

\subsection{Experimental Setup}
\label{subsec:setup}

\paragraph{Dataset and Evaluation Protocols}
We use the men's-singles records of the Match Charting Project~\citep{sackmannMCP}, containing 4.62 million valid shot events from 5,568 matches involving 705 players (2010--2026). All partitions are match-disjoint, and predictions use only the pre-contact state and preceding events within the current point. The preliminary study uses 3,898/835/835 training/validation/test matches.
\paragraph{Baselines and Implementation}
We compare \textit{HITinnes} with count-based, neural, and alternative multi-attribute predictors. Count-based controls use global or state frequencies and first- or second-order Markov models. Neural controls include a multilayer perceptron (MLP), bidirectional long short-term memory (BiLSTM)~\citep{hochreiter1997lstm,schuster1997bidirectional}, a convolutional neural network followed by long short-term memory (CNN--LSTM), and Transformer~\citep{vaswani2017attention}. Output modeling alternatives use independent heads and flat joint classification. The archived Geometry--J1 reference uses the same enriched inputs and Transformer encoder as the neural control but adopts a conditional decoder and a different checkpoint-selection objective.

\textit{HITinnes} employs a two-layer causal Transformer ($d_e=64$, four attention heads, feed-forward width 192) and a context-fusion network ($d_h=128$), with dropout 0.15 and a four-contact history window. We train using AdamW (learning rate $3\times10^{-4}$, weight decay $10^{-4}$, batch size 1,024) for up to 20 epochs with early stopping (patience 3). Matched comparisons retain the input representation, encoder, and training budget, while varying the components under study. Repeated experiments use three random seeds.

\paragraph{Evaluation Metrics}
Our primary metric is Macro-NLL, which equally weights the six phase-specific prediction targets:
\begin{equation}
\operatorname{Macro\text{-}NLL}
=\frac{1}{6}\sum_{j=1}^{6}
\frac{-\sum_t m_{t,j}\log\bar p_{t,j}(a_{t,j})}
{\sum_t m_{t,j}},
\label{eq:macro-nll}
\end{equation}
where $\bar p_{t,j}$ denotes the predicted target marginal and $m_{t,j}$ indicates an applicable, annotated target. We also report Macro-F1, averaged over classes within each target and then across the six targets. Evaluation masks are independent of training prerequisite masks. Repeated results are reported as mean $\pm$ standard deviation.

\subsection{Overall Prediction Performance}
\label{subsec:rq-estimator}

Table~\ref{tab:overall-performance} summarizes count-based predictors, neural sequence predictors, and output modeling alternatives. Within the generic-predictor block, learned sequence representations improve likelihood over count-based controls, while the evaluated neural backbones have similar scores.

Within the matched output-model block, \textit{HITinnes} obtains the lowest Macro-NLL and highest Macro-F1. Independent Heads uses the same history representation and encoder architecture, so this comparison evaluates the conditional output design under controlled inputs. Flat Joint assigns separate scores to complete tuples. \textit{HITinnes} shares conditional prediction components across action combinations. The following analyses examine representation, supervision, and inference choices.

\begin{table}[t]
\centering
\small
\setlength{\tabcolsep}{3pt}
\renewcommand{\arraystretch}{1.06}
\begin{tabular}{@{}lcc@{}}
\toprule
Method & Macro-NLL $\downarrow$ & Macro-F1 $\uparrow$\\
\midrule
\multicolumn{3}{@{}l}{\emph{Count-based predictors}}\\
Global frequency
& 1.1260 & 0.1525\\
State frequency
& 1.1038 & 0.1802\\
First-order Markov
& 1.0590 & 0.2277\\
Second-order Markov
& 1.0570 & 0.2304\\
\midrule
\multicolumn{3}{@{}l}{\emph{Neural sequence predictors}}\\
MLP
& $0.9170\pm0.0009$
& $0.2912\pm0.0012$\\
BiLSTM
& $0.9170\pm0.0003$
& $0.2902\pm0.0009$\\
CNN--LSTM
& $0.9176\pm0.0007$
& $0.2907\pm0.0011$\\
Transformer
& $0.9158\pm0.0008$
& $0.2941\pm0.0008$\\
Geometry--J1
& $0.9154\pm0.0006$
& $0.2945\pm0.0008$\\
\midrule
\multicolumn{3}{@{}l}{\emph{Output modeling alternatives}}\\
Independent heads
& $0.9203\pm0.0030$
& $0.2921\pm0.0022$\\
Flat joint
& $0.9472\pm0.0040$
& $0.2850\pm0.0027$\\
\textbf{\textit{HITinnes}}
& $\mathbf{0.8931\pm0.0030}$
& $\mathbf{0.3021\pm0.0024}$\\
\bottomrule
\end{tabular}
\caption{Overall prediction performance.}
\label{tab:overall-performance}
\end{table}

Geometry--J1 and the standard Transformer show comparable performance under the enriched-input protocol. Their mean Macro-NLL differs by approximately 0.0004 nats, compared with across-seed SDs of 0.0006 and 0.0008. The two configurations share the enriched inputs and encoder and differ in their decoder and checkpoint objective.

\subsection{Analysis of Modeling Choices}
\label{subsec:rq-design}

\begin{figure}[t]
\centering
\begin{subfigure}[t]{0.45\columnwidth}
\includegraphics[width=\linewidth]{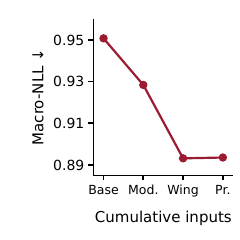}
\caption{Macro-NLL.}
\end{subfigure}\hfill
\begin{subfigure}[t]{0.45\columnwidth}
\includegraphics[width=\linewidth]{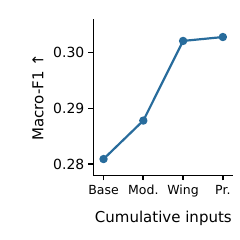}
\caption{Macro-F1.}
\end{subfigure}
\caption{Representation enrichment.}
\label{fig:understanding-hitinnes}
\end{figure}

\paragraph{Interaction-Aware Representation}
Figure~\ref{fig:understanding-hitinnes}(a) tracks Macro-NLL across cumulative input configurations. Adding categorical modifiers reduces NLL from 0.9508 to 0.9284. Receiver-relative target-wing encoding yields the largest further reduction, to 0.8931. Pressure context changes NLL to 0.8935, so the target-wing configuration remains the likelihood reference. Figure~\ref{fig:understanding-hitinnes}(b) shows the corresponding Macro-F1 values of 0.2809, 0.2878, 0.3021, and 0.3028. Pressure thus provides a small classification gain while slightly increasing likelihood loss. Base, Mod., Wing, and Pr. denote basic history and the cumulative additions of modifiers, target wing, and pressure, respectively.

The target-wing gain is consistent with interaction-relative semantics. The same charted direction can target different receiving wings depending on handedness. A separate three-seed fixed-decoder study supports the enriched input bundle, reducing Macro-NLL from $0.9773\pm0.0004$ to $0.9154\pm0.0006$ and increasing Macro-F1 from 0.2745 to 0.2945. Its largest paired target-NLL reductions occur for return stroke ($0.1926\pm0.0014$) and rally stroke ($0.1578\pm0.0013$). This repeated input comparison uses a separate protocol and changes multiple features together.

\paragraph{Partial-Label Learning and Marginal Inference}
We examine the effects of partial-label supervision and exact marginal inference in Table~\ref{tab:supervision-inference}. Both ablations increase Macro-NLL, indicating the benefit of the complete learning and inference configuration.

\textit{Partial-label supervision.}
Incomplete action annotations limit complete-tuple training but can still provide supervision for individual conditional factors. Among 73,273 return contacts in the training partition, 50,575 have complete return-action labels. Factor-wise supervision retains 22,492 additional stroke labels and 19,231 additional direction labels beyond complete-tuple supervision. This additional training signal helps explain the improvement over the complete-tuple variant, which retains the same loss formulation and training settings.

\textit{Marginal inference.}
The greedy variant reuses the full model checkpoint and conditions downstream predictions on the most likely upstream attributes. Exact marginalization averages over the possible upstream outcomes to obtain each target distribution. The lower Macro-NLL of exact marginalization supports propagating uncertainty across conditional factors.

\begin{table}[t]
\centering
\small
\setlength{\tabcolsep}{4pt}
\renewcommand{\arraystretch}{1.05}
\begin{tabular}{@{}lcc@{}}
\toprule
Configuration & Macro-NLL $\downarrow$ & Macro-F1 $\uparrow$\\
\midrule
Independent heads
& $0.9203\pm0.0030$
& $0.2921\pm0.0022$\\
w/o partial labels
& $0.9032\pm0.0032$
& $0.2980\pm0.0023$\\
Greedy inference
& $0.8993\pm0.0028$
& $0.3017\pm0.0021$\\
\textbf{Full \textit{HITinnes}}
& $\mathbf{0.8931\pm0.0030}$
& $\mathbf{0.3021\pm0.0024}$\\
\bottomrule
\end{tabular}
\caption{Ablation of supervision and inference strategies.}
\label{tab:supervision-inference}
\end{table}

\begin{figure}[t]
\centering
\includegraphics[width=0.90\columnwidth]{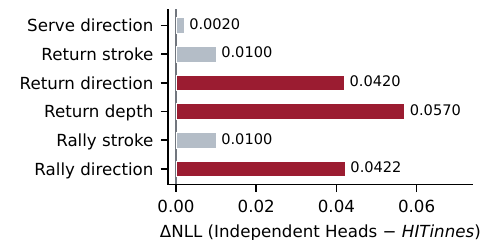}
\caption{Target-wise gains of \textit{HITinnes}.}
\label{fig:target-wise-gains}
\end{figure}

\paragraph{Target-wise Prediction Gains}
Figure~\ref{fig:target-wise-gains} shows the target-specific difference
$\Delta\operatorname{NLL}_j=\operatorname{NLL}_{\mathrm{Ind.},j}-\operatorname{NLL}_{\textit{HITinnes},j}$.
Positive values favor \textit{HITinnes}. The largest gain is for return depth (0.0570), followed by rally direction (0.0422) and return direction (0.0420). Return and rally stroke each improve by 0.0100, while serve direction changes by 0.0020. These six differences average to 0.0272 nats, matching the aggregate difference between independent heads and \textit{HITinnes} in Table~\ref{tab:overall-performance}.

The strongest gains occur for attributes modeled through downstream conditional factors, consistent with the intended role of multi-attribute prediction. Serve direction is predicted directly from the context representation and shows the smallest difference. Red bars identify downstream conditional targets and gray bars identify directly predicted targets. The representation and target-wise comparisons report point estimates.

\section{Conclusion}
We investigated history-conditioned multi-attribute prediction in tennis shot sequences. Controlled context analysis identified the predictive value of recent interactions and motivated a compact history window. Building on this finding, \textit{HITinnes} combines interaction-aware history representation with phase-dependent conditional prediction, partial-label learning, and exact marginal inference. Empirical comparisons support the effectiveness of this formulation, while component studies further characterize the roles of input representation and probabilistic estimation.
The findings characterize observed shot behavior in professional men's singles records, providing a basis for extending structured event prediction to richer observations and broader player populations.

\bibliography{references}

\end{document}